\documentclass[pdflatex,sn-mathphys-num]{sn-jnl}% Math and Physical Sciences Numbered Reference Style
\usepackage{graphicx}%
\usepackage{multirow}%
\usepackage{amsmath,amssymb,amsfonts}%
\usepackage{amsthm}%
\usepackage{mathrsfs}%
\usepackage[title]{appendix}%
\usepackage{xcolor}%
\usepackage{textcomp}%
\usepackage{manyfoot}%
\usepackage{booktabs}%
\usepackage{algorithm}%
\usepackage{algorithmicx}%
\usepackage{algpseudocode}%
\usepackage{listings}%
\theoremstyle{thmstyleone}%
\theoremstyle{thmstyletwo}%

\theoremstyle{thmstylethree}%

\begin{document}

\title[Article Title]{FlowCast-ODE: Continuous Hourly Weather Forecasting with Dynamic Flow Matching and ODE Solver}


\author[1]{\fnm{Shuangshuang} \sur{He}}\email{heshuangshuang816@gmail.com}

\author*[1]{\fnm{Yuanting} \sur{Zhang}}\email{zhangyuanting0925@gmail.com}

\author[1]{\fnm{Hongli} \sur{Liang}}\email{helenaliang9@gmail.com}

\author[1]{\fnm{Qingye} \sur{Meng}}\email{hilbertmeng@gmail.com}

\author[1]{\fnm{Xingyuan} \sur{Yuan}}\email{yuan@caiyunapp.com}

\author[2,3]{\fnm{Shuo} \sur{Wang}}\email{shuowang.ai@gmail.com}

\affil[1]{\orgname{ColorfulClouds Technology Co.,Ltd.}, \orgaddress{\city{Beijing}, \country{China}}}

\affil[2]{\orgname{School of Systems Science, Beijing Normal University}, \orgaddress{\city{Beijing}, \country{China}}}

\affil[3]{\orgname{D-ITET, ETH Zurich}, \orgaddress{\city{Zurich}, \country{Switzerland}}}



\abstract{Data-driven hourly weather forecasting models often face the challenge of error accumulation in long-term predictions. The problem is exacerbated by non-physical temporal discontinuities present in widely-used training datasets such as ECMWF Reanalysis v5 (ERA5), which stem from its 12-hour assimilation cycle. Such artifacts lead hourly autoregressive models to learn spurious dynamics and rapidly accumulate errors. To address this, we introduce FlowCast-ODE, a novel framework that treats atmospheric evolution as a continuous flow to ensure temporal coherence. Our method employs dynamic flow matching to learn the instantaneous velocity field from data and an ordinary differential equation (ODE) solver to generate smooth and temporally continuous hourly predictions. By pre-training on 6-hour intervals to sidestep data discontinuities and fine-tuning on hourly data, FlowCast-ODE produces seamless forecasts for up to 120 hours with a single lightweight model. It achieves competitive or superior skill on key meteorological variables compared to baseline models, preserves fine-grained spatial details, and demonstrates strong performance in forecasting extreme events, such as tropical cyclone tracks.}

\keywords{temporal continuity, hourly weather forecasting, dynamic flow matching, ode solver}



\maketitle

\section{Introduction}
\label{sec:intro}

Accurate weather forecasting is a cornerstone of modern society, critical for sectors ranging from energy management to disaster preparedness \cite{bauer2015quiet}. The field has long been dominated by traditional numerical weather prediction (NWP) systems, which simulate atmospheric physics with high fidelity but at a significant computational cost \cite{Ritchie1995, Wedi2014}. Recently, a paradigm shift has occurred with the advent of data-driven models like Pangu-Weather, GraphCast, FourCastNet \cite{biAccurateMediumrangeGlobal2023, lam2023learning, pathak2022fourcastnet}, among others \cite{chen2023b, fengwu, nguyen2023scaling,lang2024, aurora}. Trained on decades of reanalysis data such as ECMWF Reanalysis v5 (ERA5) \cite{hersbach2020era5}, these models have achieved forecasting skill comparable or even superior to state-of-the-art NWP systems at medium ranges, while reducing inference times by orders of magnitude \cite{Bouallegue2024Rise, Ling2024Improving}. This success, however, has primarily been demonstrated at 6-hour temporal resolution, masking a fundamental challenge that emerges at higher frequencies.

\begin{figure}[!t]
  \centering
  \includegraphics[width=0.98\textwidth]{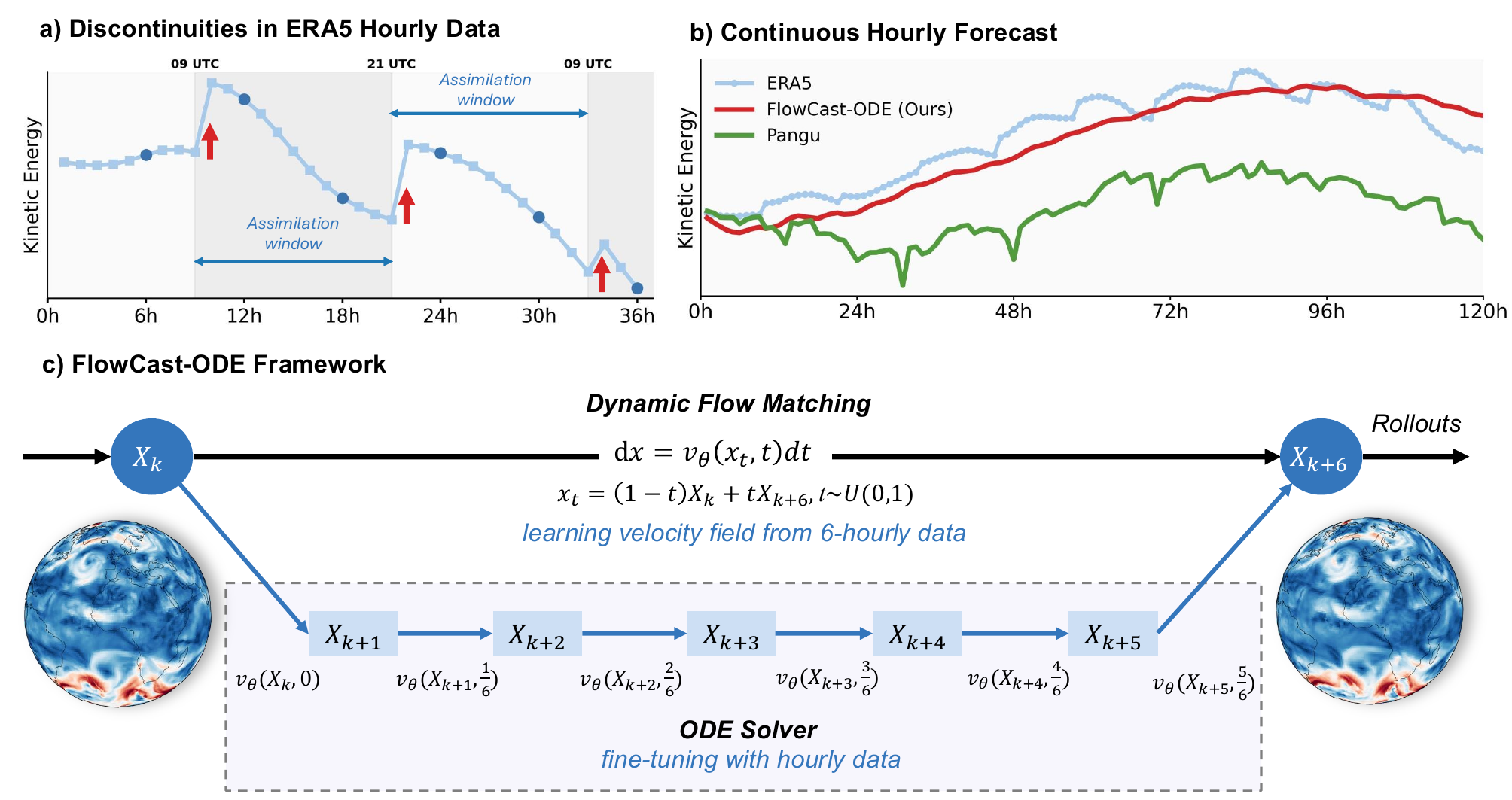}
  \caption{\textbf{FlowCast-ODE: A continuous-time framework for stable hourly weather forecasting.} This figure details FlowCast-ODE's motivation, performance, and methodology. \textbf{a} Discontinuities in ERA5 Hourly Data. ERA5, a foundational training dataset for data-driven weather models, exhibits systemic, non-physical temporal jumps (e.g., in kinetic energy) at the 12-hour data assimilation boundaries (09 and 21 UTC). These artifacts mislead hourly autoregressive models, causing inconsistent physical dynamics and cumulative forecast errors. Most data-driven models, operating on 6-hour intervals (deep blue dots), often unintentionally bypass this critical discontinuity issue. \textbf{b} Continuous Hourly Forecast. A 120-hour forecast of kinetic energy demonstrates that FlowCast-ODE (red, ours) produces remarkably smooth and temporally coherent predictions, outperforming both the fluctuating Pangu-Weather baseline (green) and the original ERA5 data (light blue). \textbf{c} FlowCast-ODE Framework. A two-stage training process to achieve temporally continuous hourly weather forecasting on discontinuous data. First, the velocity field, $v_\theta(x_t, t)$, is learned on 6-hour intervals using a linearized Dynamic Transport path, to continuously transform an initial atmospheric state $X_k$ into target $X_{k+6}$. By training on 6-hour interval data, FlowCast-ODE inherently reduces the impact of these discontinuities. Subsequently, the model is finetuned on hourly data, where an Euler ODE solver integrates the learned dynamics in hourly steps $(X_k \to X_{k+1}, \ldots, X_{k+6})$, yielding stable hourly predictions. For long-range prediction, the solver performs an hourly rollout within 6-hour intervals, with the final state of each interval serving as the initial condition for the next, thus extending the forecast horizon autoregressively.}
\label{fig:method-overview}
\end{figure}

The push toward hourly forecasting, essential for capturing rapidly evolving phenomena \cite{Pielke2002Weather}, reveals a systemic challenge beyond error accumulation \cite{nguyen2023}: limitations within the foundational data induce inconsistent physical dynamics in the models trained upon it. The ERA5 reanalysis dataset, despite its widespread use, contains systemic, non-physical temporal discontinuities originating from its 12-hour data assimilation cycle \cite{yuan2025}. This process introduces artificial jumps in atmospheric state around 09:00 and 21:00 UTC (see Fig.~\ref{fig:method-overview}a and b). When autoregressive models are trained on hourly data, they are forced to learn these spurious dynamics. Consequently, their internal representation of atmospheric physics becomes corrupted, leading to a rapid accumulation of forecast errors, a problem that cannot be solved by simply increasing model capacity or refining architectures \cite{schreck2024, moldovan2025updateecmwfsmachinelearnedweather}. Existing hourly models either avoid this issue with complex multi-model stitching \cite{biAccurateMediumrangeGlobal2023}, creating inconsistent forecast trajectories, or integrate physical cores that sacrifice the computational efficiency of pure data-driven approaches \cite{neuralgcm}.

This challenge motivates a return to the physical principles of atmospheric dynamics. The evolution of the atmosphere is a continuous process governed by differential equations; therefore, an ideal forecasting model should embody this continuity in its design, predicting future states from known initial conditions. The task of learning such complex, high-dimensional transformations that represent this evolution falls squarely within the domain of modern generative modeling, for which flow matching offers a highly promising framework. Flow matching trains a neural network to learn a conditional velocity field that defines an Ordinary Differential Equation (ODE), which in turn smoothly transports samples from a prior distribution to a target data distribution \cite{Lipman2023, holderrieth2025generative}. 
Flow matching is better suited for atmospheric modeling than another mainstream generative model, diffusion models \cite{ho2020denoising, song2021denoising, song2022denoising}, as it is formulated on continuous normalizing flows, which align with the continuous physical processes of atmospheric dynamics, and it learns an interpretable velocity field. However, conventional flow matching often share a conceptual starting point with diffusion: learning a mapping from a Gaussian prior to the data distribution. We argue that a more physically coherent approach is to learn a dynamic conditional flow that evolves the atmospheric state directly from the previous time step. This reframes the task from a "noise-to-data" generation problem into a dynamic "data-to-data" evolution problem \cite{liu2023flow, lim2025a}. Such a shift not only greatly simplifies the learning task but also focuses it on the more physically meaningful process of short-term evolution.

Based on these physical and methodological considerations, we introduce FlowCast-ODE, a continuous-time framework designed specifically for stable and temporally coherent hourly weather forecasting. At the core of our method, we adapt flow matching to learn a conditional flow path directly from the previous state, rather than relying on a Gaussian noise field \cite{lim2025a}, making it ideally suited for deterministic forecasting from known initial conditions. However, to ensure this model is robust and not corrupted by data artifacts, we introduce a principled two-stage training strategy. Recognizing the challenge of learning directly from imperfect hourly data, the model is first pre-trained on coarse, 6-hour intervals using a linearized dynamic flow path. This simplification acts as a powerful regularizer, forcing the model to first learn a stable and foundational velocity field that captures the main 6-hour dynamics while ignoring assimilation jumps. Subsequently, this robust base model is fine-tuned on hourly data, where the ODE solver integrates the learned velocity field to resolve the high-frequency, non-linear variations and produce a continuous forecast trajectory. This approach ensures that the learned dynamics are both physically sound and temporally coherent.

This paper makes the following key contributions:

\begin{itemize}
    \item Revealing inherent discontinuities in reanalysis data.
    By analyzing atmospheric kinetic energy, we quantify systemic temporal discontinuities in the foundational ERA5 dataset, as illustrated in Fig.~\ref{fig:method-overview}a. These non-physical “jumps,” caused by the 12-hour assimilation cycle, lead to rapid error accumulation in hourly iterative models. Consequently, models trained directly on this data inevitably learn spurious physical dynamics at assimilation boundaries.

    \item A novel framework for continuous-time forecasting.
    Grounded in the physical principle that atmospheric evolution is continuous, we propose \textbf{FlowCast-ODE}, which learns the velocity field governing the evolution from an initial state to a future state. A two-stage training strategy ensures robust learning from imperfect data. As depicted in Fig.~\ref{fig:method-overview}c, the framework first learns the foundational velocity field from linearized dynamic flow path at 6-hour intervals. Subsequently, an ODE solver integrates this field during fine-tuning stage to produce smooth, temporally coherent forecast trajectories. 

    \item A lightweight model with superior performance.
    FlowCast-ODE delivers a purely data-driven, hourly iterative model with notable advantages:
    \begin{itemize}
        \item Accuracy: Achieves competitive or superior skill on key meteorological variables compared to leading models, preserves fine-grained spatial details, and demonstrates strong skill in extreme events, such as tropical cyclone tracks.
        \item Continuity: Generates forecasts that maintain physical temporal continuity across assimilation window boundaries (see Fig.~\ref{fig:method-overview}b).
        \item Efficiency: Incorporates temporal conditioning via techniques such as low-rank decomposition, enabling hourly forecasts of up to 120-hours with a single lightweight model, eliminating the need for complex multi-model stitching approaches.
    \end{itemize}
\end{itemize}

\section{Results}
\label{sec:results}
\subsection{Baseline models}

Most existing AI-based weather models are designed for 6-hour forecast intervals, whereas Pangu-Weather \cite{biAccurateMediumrangeGlobal2023} addresses this by training separate models for 1-, 3-, 6-, and 24-hour forecast horizons, enabling hourly forecasts and controlling error growth over longer lead times. Its accuracy and robustness under consistent datasets and training configurations \cite{Gao2025} make it a strong baseline for evaluating our proposed method. We also include ClimODE \cite{verma2024climode}, a physics-informed model that represents weather as a continuous-time spatiotemporal process with advection-driven, value-conserving neural flows. However, due to ClimODE running significantly slower when increasing resolution from 5.625° to 1° and is designed for only five variables, we rely on results reported in its paper. We adopt both Pangu-Weather and ClimODE as representative baselines to assess our approach. We run forecasts for FlowCast-ODE and Pangu-Weather using ERA5 reanalysis as initial conditions. Forecasts are initialized at 00Z and 12Z from 1 January to 31 December 2021, with lead times up to 120 hours. To compare error growth over long lead times, Pangu-Weather employs only its 24-hour model.

\subsection{Skill}
\begin{figure}[h!t]
\centering
\includegraphics[width=0.98\textwidth]{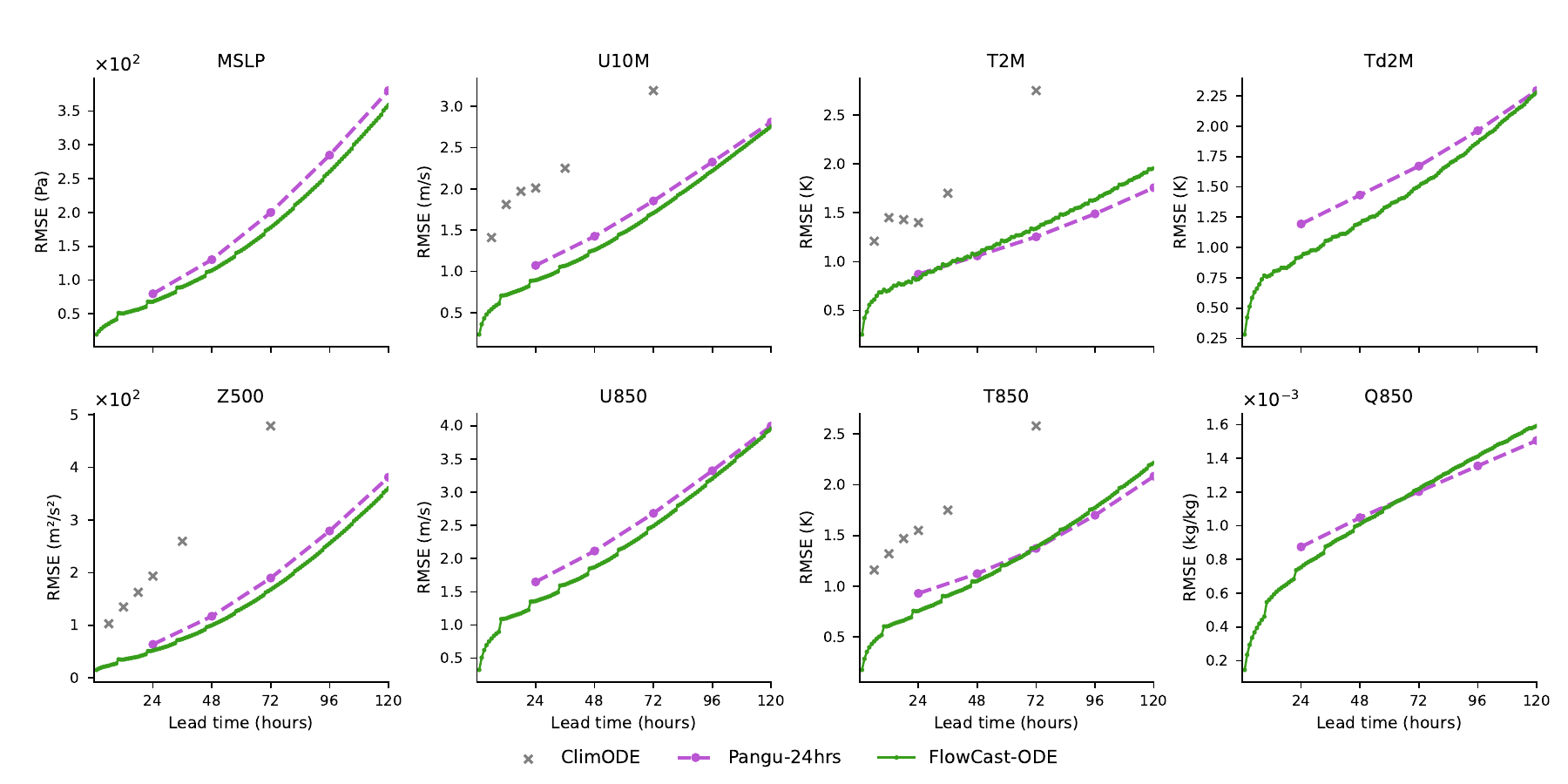}
\caption{\textbf{Comparison of latitude-weighted global mean RMSE of ClimODE (gray x markers), Pangu-Weather (24-hour intervals, purple dashed line), and FlowCast-ODE (hourly resolution, green line) across a 5-day forecast horizon.} 
Results are shown for four surface variables (MSLP, U10M, T2M, and TD2M) and four upper-air variables (Z500, U850, T850, and Q850). 
The hourly RMSE of FlowCast-ODE reveals faster error growth around 09:00–10:00 UTC and 21:00–22:00 UTC, reflecting the impact of ERA5’s discontinuities at the boundaries of the 12-hour assimilation windows on the hourly model’s forecast accuracy.}
\label{fig:rmse}
\end{figure}

We compare long-horizon forecast performance for FlowCast-ODE and Pangu-Weather in terms of root mean square error (RMSE). Fig.~\ref{fig:rmse} shows the latitude-weighted global mean RMSE for four surface variables (MSLP, U10M, T2M, TD2M) and four upper-air variables (Z500, U850, T850, Q850) over a 5-day forecast horizon. FlowCast-ODE achieves lower RMSE for MSLP and Z500 across all lead times. For U10M, U850, and TD2M, it outperforms Pangu-Weather at shorter lead times, with the margin decreasing as the forecast horizon increases. For Q850 and T850, FlowCast-ODE maintains an advantage within the first 72 hours, after which Pangu-Weather slightly surpasses it. For T2M, FlowCast-ODE shows superior skill during the first 24 hours, while Pangu-Weather performs better at longer lead times.

Specifically, for the 120-hour lead time, Pangu-Weather generates forecasts by iteratively applying its 24-hour model five times, whereas FlowCast-ODE produces hourly forecasts through 120 iterations. Despite the larger number of iterations, FlowCast-ODE maintains competitive, and often superior RMSE performance, maintaining stability and low error across multiple iterations and long-horizon forecasts.

\subsection{Power Spectra}
\label{sec:text_spectra}
\begin{figure}[!t]
  \begin{center}
    \includegraphics[width=0.99\textwidth]{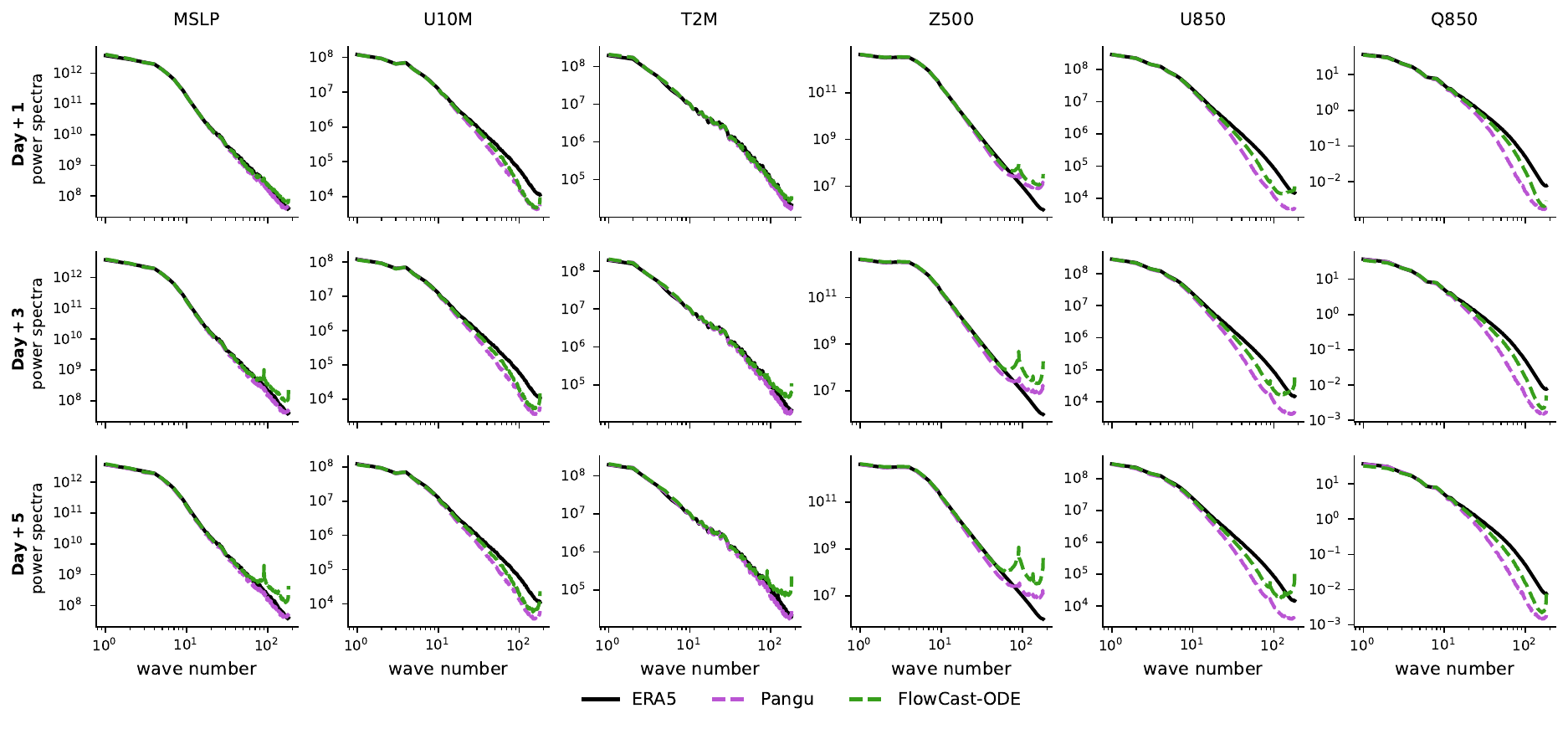}
    \caption{\textbf{Latitude-averaged power spectra from 60$^\circ$S to 60$^\circ$N for ERA5 (black line), Pangu-Weather forecasts (purple dashed line), and FlowCast-ODE forecasts (green dashed line).} Rows 1, 2, and 3 correspond to forecast days 1, 3, and 5, respectively.}
    \label{fig:spectrum}
  \end{center}
\end{figure}

While RMSE provides a global measure of forecast accuracy, it does not capture how well models preserve spatial structures across different scales. To evaluate this aspect, we analyze the latitude-averaged (60°N–60°S) power spectra of six variables in ERA5 for the year 2021, along with predictions from Pangu-Weather and FlowCast-ODE (Fig.~\ref{fig:spectrum}). For U10, U850, and Q850, both FlowCast-ODE and Pangu-Weather underestimate power at short-to-mid wavelengths. Nonetheless, FlowCast-ODE consistently retains higher spectral energy, indicating a stronger capability to preserve fine-scale variability. For MSLP, Z500, and T2M, FlowCast-ODE slightly overestimates short-wave energy, producing a spurious peak near wavenumber 90 that grows with lead time. As this wavenumber corresponds to the training patch size, we tentatively attribute this artifact to it and aim to address it in future work.

\begin{figure}[!t]
  \begin{center}
    \includegraphics[width=0.98\textwidth]{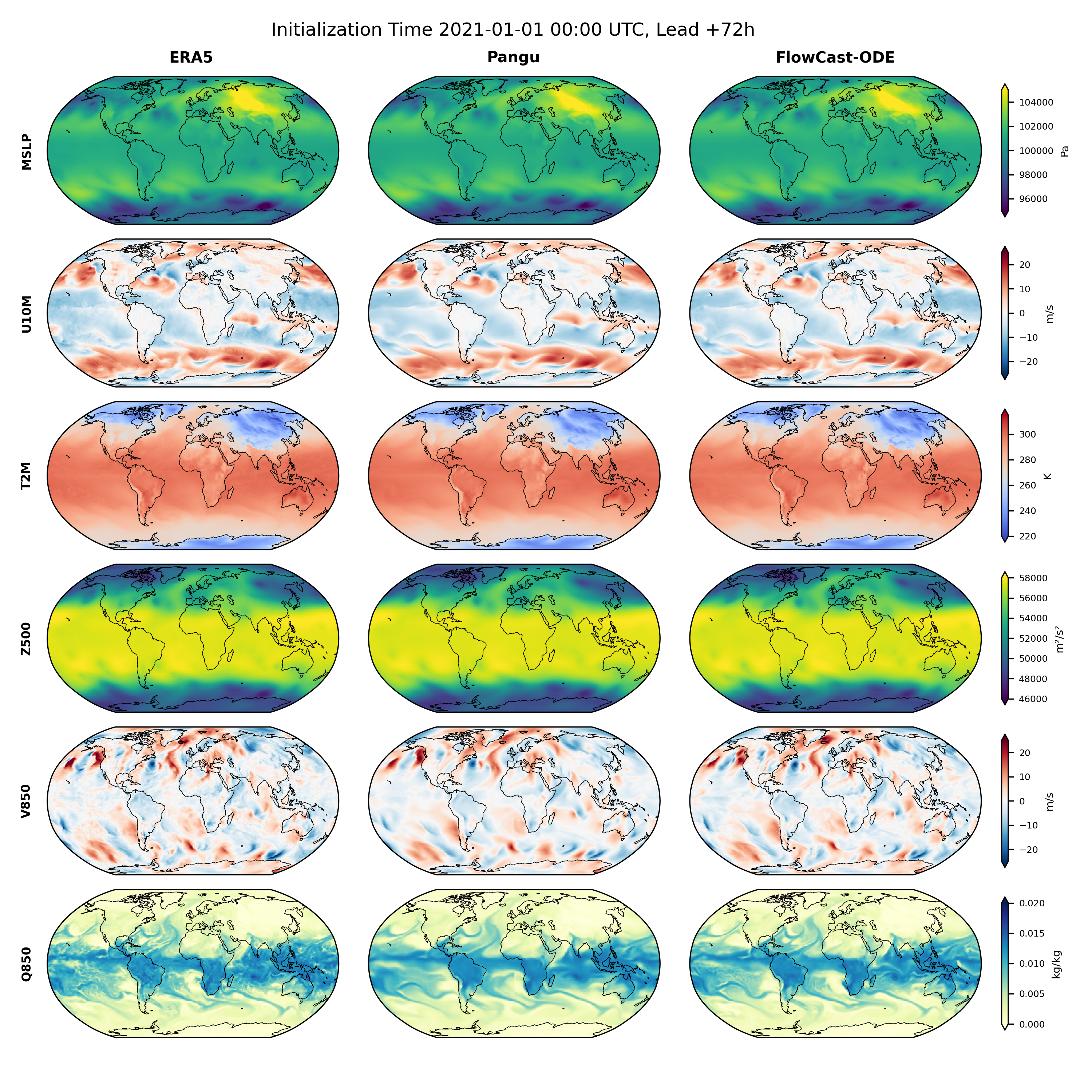}
    \caption{\textbf{Visualization of 3-day forecast results for three surface variables (MSLP, U10M, T2M) and three upper-air variables (Z500, V850, Q850).} For each variable, ERA5 ground truth (left), Pangu-Weather forecast (middle), and FlowCast-ODE forecast (right) are shown. Forecasts are initialized at 00:00 UTC on 1 January 2021.}
    \label{fig:spatial-0101-72}
  \end{center}
\end{figure}

The better preservation of small-scale spectral energy by FlowCast-ODE is also observed in the spatial patterns of the forecasts. As shown in Fig.~\ref{fig:spatial-0101-72}, FlowCast-ODE effectively reduces over-smoothing and delivers improved overall forecast accuracy compared to Pangu-Weather. Both models reproduce the large-scale structures of ERA5 in 3-day forecasts. 

\subsection{Tracking tropical cyclones}

We evaluated the track forecasting performance of six representative tropical cyclones that occurred in 2021: SURIGAE, IN-FA, MINDULLE, CHANTHU, KOMPASU, and RAI. For each cyclone, predicted centers were determined by locating the MSLP minimum based on the tracking algorithm \cite{white2005} described in Supplementary Notes 1.3. Observed tracks were obtained from the International Best Track Archive for Climate Stewardship (IBTrACS) \cite{Knapp2010, Knapp2018}, which provides the most reliable reference for tropical cyclones. To provide a strong baseline, we included forecasts from the 0.25° Pangu-Weather model downloaded from WeatherBench2.

\begin{figure}[!t]
\centering
\includegraphics[width=0.98\textwidth]{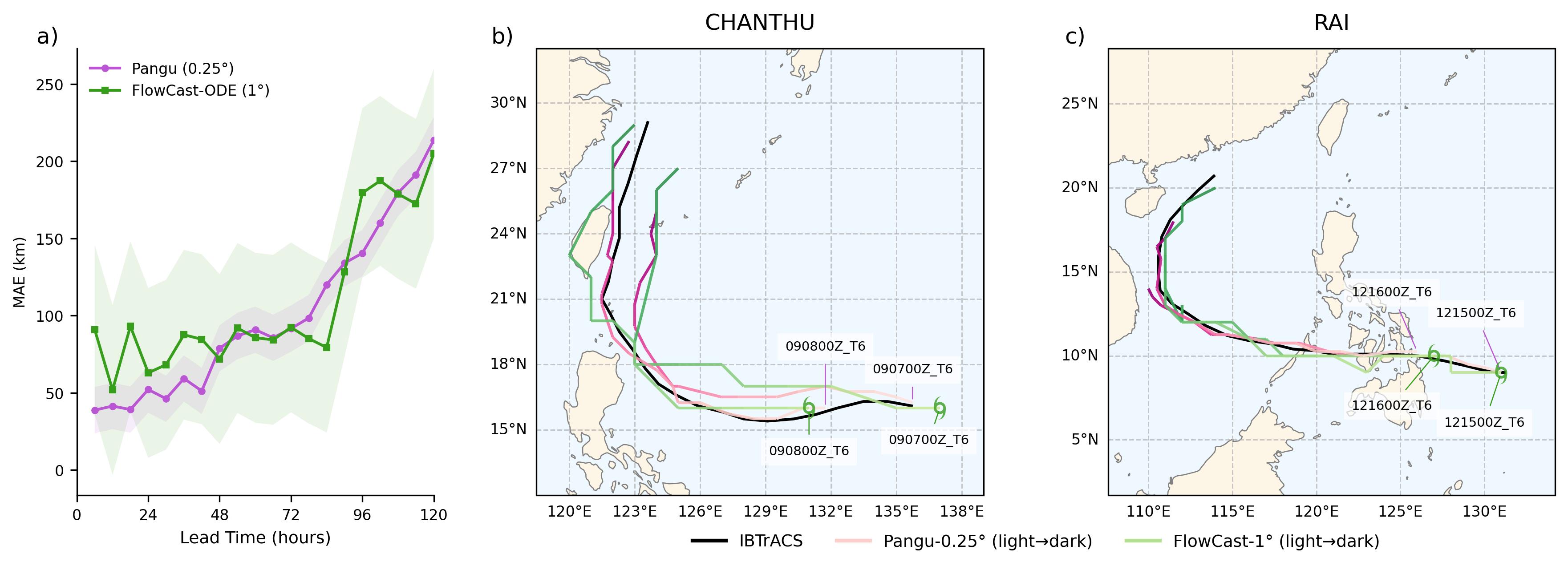}
\caption{
\textbf{Comparison of tropical cyclone track forecasts in 2021 from FlowCast-ODE and 0.25° Pangu-Weather.} 
\textbf{a} Comparison of the MAE for six tropical cyclone track forecasts over a 120-hour forecast period for both FlowCast-ODE and the 0.25° Pangu-Weather. The shaded area indicates the spatial extent represented by each model’s grid. 
\textbf{b, c} Tracking results for Typhoon CHANTHU and RAI. Observed IBTrACS tracks (black line) and predicted trajectories at 6-hour intervals from FlowCast-ODE 1° (green line) and 0.25° Pangu-Weather (purple line). Typhoon markers denote positions six hours after forecast initialization, and initialization times are annotated in the panels. Line colors transition from light to dark with increasing lead time.}
\label{fig:typhoon_track}
\end{figure}

Track errors were quantified using the mean absolute error (MAE), representing the distance between observed and predicted cyclone centers. Fig.~\ref{fig:typhoon_track}a shows the MAE over the 120-hour forecast period for both FlowCast-ODE and Pangu-Weather. FlowCast-ODE generally exhibits slightly higher MAE than the 0.25° Pangu-Weather model within 48 hours, but its MAE becomes comparable to Pangu-Weather beyond 48 hours. Although FlowCast-ODE operates at a coarser 1° resolution compared to the 0.25° Pangu-Weather model, it demonstrates competitive performance in tropical cyclone track forecasting.
Fig.~\ref{fig:typhoon_track}b and c presents the observed IBTrACS tracks and corresponding five-day forecasts from FlowCast-ODE and Pangu-Weather across multiple initializations for typhoon CHANTHU and RAI. Both models reproduce the overall trajectory patterns.

\subsection{Assimilation-Induced Temporal Discontinuity of Atmospheric Energy}
\label{sec:tem_continuity}
\begin{figure}[!t]
  \centering
  \includegraphics[width=0.98\textwidth]{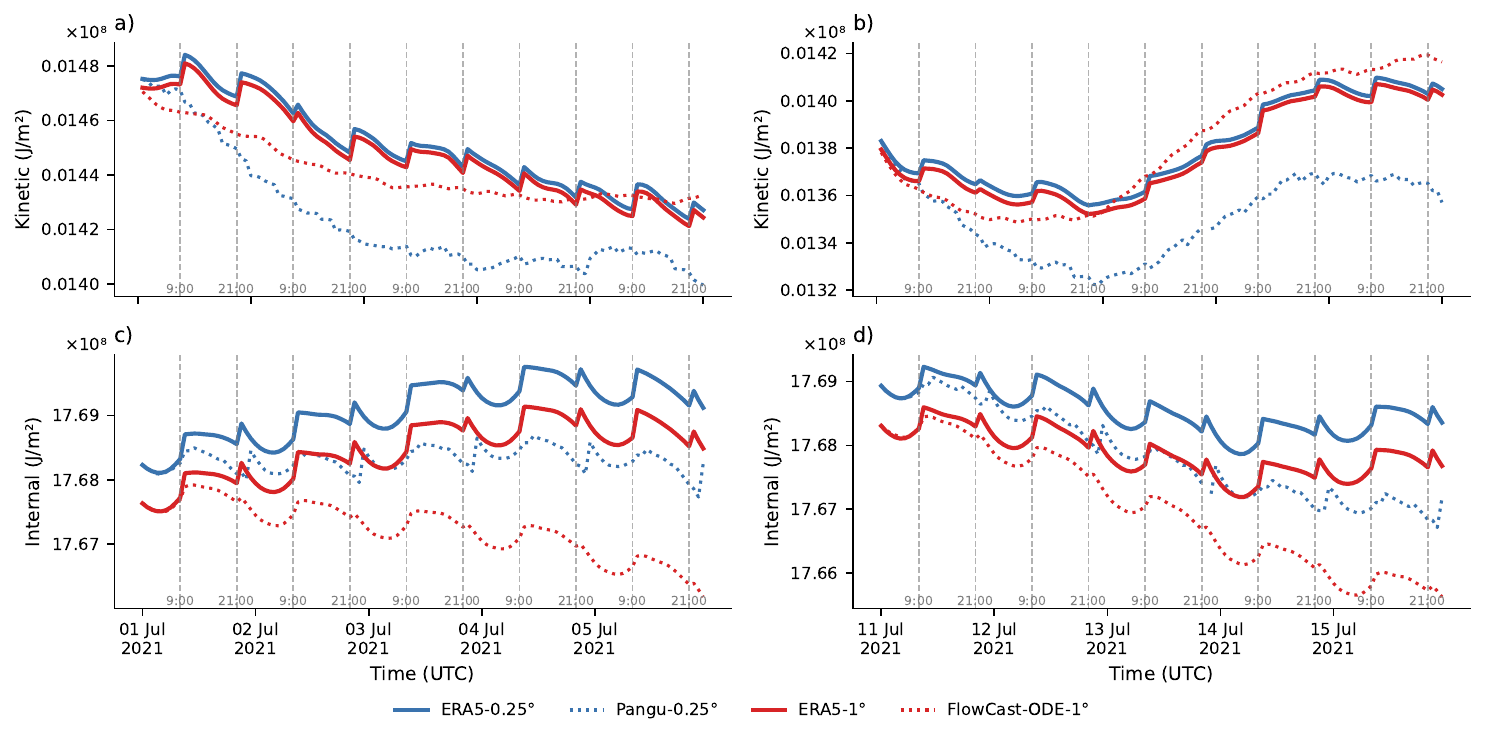}
  \caption{\textbf{Time series of kinetic energy (a, b) and internal energy (c, d) over a 5-day period.} Forecasts are initialized at 1 July 2021 (\textbf{a, c}) and 11 July 2021 (\textbf{b, d}). Depicted are ERA5 at 0.25° (solid blue) and 1° resolution (solid red), along with forecasts from Pangu-Weather at 0.25° (dashed blue) and FlowCast-ODE at 1° (dashed red). ERA5 shows spurious jumps at the boundaries of 12-hour assimilation windows, while FlowCast-ODE exhibits smoother temporal trajectory and Pangu-Weather forecasts display fluctuations.}
  \label{fig:energy_k_0721}
  \vspace{-5pt}
\end{figure}

We examine the temporal evolution of kinetic energy and internal energy using cases initialized at 00 UTC on 1 July 2021 and 00 UTC on 11 July 2021. Fig.~\ref{fig:energy_k_0721} shows the 5-day time series of ERA5 at 0.25° and 1° resolution, together with forecasts from Pangu-Weather at 0.25° (produced by the officially released model) and FlowCast-ODE at 1°.

For kinetic energy, both the 0.25° and 1° ERA5 data exhibit continuous evolution within each data assimilation window, but discontinuities appear between successive windows. FlowCast-ODE produces relatively smoother energy trajectory that closely follow the ERA5 trend, while Pangu-Weather, due to its hierarchical multi-model stitching strategy, shows fluctuations, and also a gradual underestimation of kinetic energy compared to ERA5. For internal energy, the 0.25° ERA5 values are consistently higher than those at 1°, with discontinuities also arising from assimilation cycles. Both FlowCast-ODE and Pangu-Weather underestimate internal energy relative to ERA5 at their respective resolutions. FlowCast-ODE again demonstrates better temporal continuity, whereas Pangu-Weather exhibits fluctuations. 

Notably, both models track the internal energy of ERA5 closely during the first assimilation window, but the discrepancies increase after each successive window, highlighting the influence of assimilation-induced discontinuities in ERA5 on model forecasts. Detailed definitions and computation procedures for kinetic and internal energy \cite{mayer2021, sha2025}, and the assimilation induced discontinuities present in ERA5 hourly data, are provided in the Supplementary Notes 1.1.

\section{Discussion}\label{sec:discussion}

In this study, we identified a fundamental obstacle to data-driven hourly weather prediction: the systemic, non-physical temporal discontinuities in widely used reanalysis datasets such as ERA5. We showed that these artifacts distort the learned dynamics of autoregressive models, leading to rapid error accumulation near assimilation window boundaries. Our proposed framework, FlowCast-ODE, addresses this challenge through the lens of continuous dynamical systems—a perspective grounded in the continuous nature of atmospheric evolution.

The results confirm the value of our approach. FlowCast-ODE produces temporally smooth and continuous forecasts, most notably by eliminating the artificial jumps in atmospheric energy present in the training dataset. Its competitive or superior RMSE scores not only demonstrate higher accuracy but also indicate improved stability. Despite requiring 120 iterative steps for a 5-day forecast, FlowCast-ODE avoids the rapid error growth common in hourly models. Moreover, the preservation of small-scale spatial details, as revealed by power spectral analysis, suggests that the model learns a more faithful representation of physical dynamics and is less prone to over-smoothing. Importantly, FlowCast-ODE achieves comparable skill in tracking tropical cyclones to a state-of-the-art model at higher resolution, underscoring its capability in representing extreme events.

While FlowCast-ODE represents a significant step forward, several limitations remain and define avenues for future work. The current resolution restricts the ability to resolve fine-scale phenomena. Future work will focus on scaling the framework to higher-resolution datasets. Although FlowCast-ODE is currently deterministic, its framework could be extended by replacing the ODE with a stochastic differential equation (SDE) formulation, enabling the generation of ensemble forecasts to quantify uncertainty \cite{albergo2023stochasticinterpolant, chen2024probabilisticforecastingstochasticinte}, a critical requirement for operational use. Finally, the spurious spectral energy peaks arising from the patch-based architecture indicate the potential of multi-scale model designs to better capture both local and large-scale structures, thereby improving long-term energy conservation.

In conclusion, FlowCast-ODE provides an effective framework for continuous hourly weather prediction. By integrating a continuous dynamics perspective of atmospheric evolution with data-driven learning, it enhances forecast stability and temporal coherence compared with existing approaches, making it a promising step toward more reliable ML-based weather forecasting systems.

\section{Methods}\label{methods}
Given a spatiotemporal atmospheric field \( X_k \in \mathbb{R}^{C \times H \times W} \) at time \( k \), where \( C \) denotes the number of weather variables, the goal is to autoregressively forecast a sequence of future states \( \{X_{k+1}, X_{k+2}, \dots, X_{k+N}\} \) at hourly resolution, up to a forecast horizon \( N \) (e.g., 120 hours). This task involves predicting the evolution of 71 weather variables, including 6 surface variables and 5 upper-air variables across 13 pressure levels. The objective is to learn a model that captures the continuous dynamics of the atmosphere, enabling accurate hourly weather forecasts, while mitigating issues such as assimilation-induced discontinuities.

\subsection{Dataset}

We conduct experiments on the WeatherBench2 \cite{Rasp2024} benchmark, which is a curated subset of the ERA5 \cite{hersbach2020era5} reanalysis dataset provided by the European Centre for Medium-Range Weather Forecasts (ECMWF). ERA5 is a global atmospheric reanalysis spanning from 1959 to the present, with a native spatial resolution of $0.25^\circ$ latitude/longitude and 1-hour temporal intervals. It is produced using ECMWF’s HRES model (cycle 42r1) within a 4D-Var data assimilation framework, which combines forecasts with observations in 12-hour assimilation windows (21:00–09:00 and 09:00–21:00). We adopt the $1^\circ$ resolution (corresponds to a grid size of $181 \times 360$) version of WeatherBench2. As summarized in Supplementary Table 1, the dataset we used contains five atmospheric variables: geopotential (Z), specific humidity (Q), temperature (T), U component of wind (U),  V component of wind (V) at 13 pressure levels (50, 100, 150, 200, 250, 300, 400, 500, 600, 700, 850, 925, and 1,000~hPa) and six surface variables: 10-m u wind component (U10M), 10-m v wind component (V10M), 2-m temperature (T2M), 2-m dewpoint temperature (TD2M), mean sea level pressure (MSLP) and total precipitation (TP), as well as topographical features including geopotential at surface, land-sea mask, and soil type. For model development, the data is split into 1980–2019 for training, 2020 for validation, and 2021 for testing.

To assess tropical cyclone track forecasts, we utilize the International Best Track Archive for Climate Stewardship (IBTrACS) dataset \cite{Knapp2010, Knapp2018}. IBTrACS integrates best-track data from international meteorological agencies, offering a comprehensive time series of cyclone-eye coordinates (latitude and longitude), wind speeds, and central pressures. In this work, we focus on six typhoons in 2021 from the West Pacific basin—SURIGAE, IN-FA, MINDULLE, CHANTHU, KOMPASU, and RAI—to evaluate forecast accuracy against observed tracks.

\subsection{Data Preprocessing}
\label{sec_datapp}
All variables were normalized prior to model training to ensure numerical stability and facilitate effective learning. Specifically, for each variable at each level, we calculated the mean and standard deviation using the training dataset spanning from 1980 to 2019. Normalization was performed by subtracting the corresponding mean and dividing by the standard deviation, resulting in zero-mean and unit-variance variables. For total precipitation, which typically exhibits a highly skewed distribution and strong imbalance, we applied a logarithmic transformation: $\hat{TP} = 10 \cdot \log_{10}\Big( 1 + 200 \cdot TP^{1.6} \Big)$.

\subsection{FlowCast-ODE Model}
\subsubsection{Model Overview}
As illustrated in Fig~\ref{fig:method-overview}c, FlowCast-ODE employs a two-stage training process to achieve temporally continuous hourly weather forecasting. This strategy is designed to first learn the fundamental atmospheric dynamics on coarse intervals and then refine the model to capture non-linear hourly variations.

In the first stage, we pre-train the model on 6-hour data intervals (00/06/12/18:00 UTC) to minimize the impact of data assimilation inconsistencies present in the full hourly ERA5 dataset. Following the flow matching framework, the velocity model $v_{\theta}$ is defined as the instantaneous time derivative of the atmospheric state.
\begin{equation}
\frac{d{x}(t)}{dt} = v_{\theta}({x}_t, t, {c}),
\end{equation}

${c}$ denote conditioning features, including land–sea mask, surface geopotential, soil type, spatial coordinates, and temporal features. 

We utilize a Dynamic Transport path (see Section~\ref{sec_dynamicfm} for details), which reframes the objective from the standard "noise-to-data" problem to a "data-to-data" evolution, aligning with the actual atmospheric evolution from its previous state. Dynamic Transport path assumes a linear evolution path between the start state and end state. Under this assumption, the target velocity reduces to $X_{k+6} - X_k$. This stage yields a robust model that captures the fundamental 6-hour atmospheric flow based on a simplified, linear assumption.

In the second stage, we fine-tune the pre-trained model using hourly data for capturing non-linear atmospheric evolution. We use an explicit Euler scheme \cite{iserles2009} to autoregressively solve the ODE at hourly resolution:

\begin{equation}
{X}_{k+n} = {X}_{k+n-1} + \delta_t \cdot v_{\theta}({X}_{k+n-1}, {t_n}, {c})
\end{equation}

where $\delta_t = \frac{1}{6}$ is the normalized time step for a 1-hour forecast, and $t_n =\frac{n-1}{6}$ ($n=1,2,\dots6$) is the normalized time at the start of the step. This defines a 6-hour prediction block, which can be sequentially applied for longer forecasts. The normalized time input $t_n$ is reset to zero at the start of each block. For generating the final forecasts, the fully fine-tuned model is applied in the same autoregressive manner to produce the hour-by-hour predictions.

Overall, by combining 6-hour interval dynamic flow matching with ODE-based refinement, this two-stage framework allows the model to first establish a stable foundation from coarse 6-hour data and then refine it to capture the high-resolution dynamics of hourly predictions, generating fine-grained, temporally coherent forecasts.

\subsubsection{Dynamic Flow Matching}
\label{sec_dynamicfm}
Flow Matching \cite{Lipman2023} is a generative modeling approach based on conditional normalizing flows \cite{dinh2014nice, Chen2018}, which learn time-dependent velocity fields along probability paths, enabling continuous transformation from a simple distribution (e.g., Gaussian) to a complex target data distribution. The core idea is to characterize the trajectory of samples \(\phi_t(x)\) over time \(t \in [0,1]\) as the solution to the ordinary differential equation:
\begin{equation}
\frac{d}{dt} \phi_t(x) = v_t(\phi_t(x)),
\label{eq:flow_dynamics}
\end{equation}
where \(v_t(\cdot)\) denotes the parameterized velocity field to be learned.

The training objective minimizes the flow matching loss, which is equivalent to the conditional flow matching (CFM) loss:
\begin{equation}
\mathcal{L}_{CFM}(\theta) = \mathbb{E}_{t, x_1, x \sim p_t(x|x_1)} \left\| v_t(x; \theta) - u_t(x|x_1) \right\|^2,
\label{eq:cfm_loss}
\end{equation}
where \(x_1\) is sampled from the target distribution. The conditional probability path is assumed Gaussian:
\begin{equation}
p_t(x|x_1) = \mathcal{N}\left(x \mid \mu_t(x_1), \sigma_t(x_1)^2 I\right),
\label{eq:pt_gaussian}
\end{equation}
with boundary conditions \(\mu_0(x_1) = 0, \sigma_0(x_1) = 1\), corresponding to a standard Gaussian, and \(\mu_1(x_1) = x_1, \sigma_1(x_1) = \sigma_{\min}\), corresponding to a narrow Gaussian around \(x_1\). The conditional flow transformation is:
\begin{equation}
\psi_t(x) = \sigma_t(x_1) x + \mu_t(x_1),
\label{eq:flow_transform}
\end{equation}
and the corresponding conditional velocity field is:
\begin{equation}
u_t(x|x_1) = \frac{\sigma_t'(x_1)}{\sigma_t(x_1)} (x - \mu_t(x_1)) + \mu_t'(x_1).
\label{eq:velocity_field}
\end{equation}

By reparameterizing the sample as \(x_0 \sim \mathcal{N}(0, I)\), the loss becomes:
\begin{equation}
\mathcal{L}_{CFM}(\theta) = \mathbb{E}_{t, q(x_1), p(x_0)} \left\| v_t(\psi_t(x_0)) - \frac{d}{dt} \psi_t(x_0) \right\|^2.
\label{eq:cfm_loss_reparam}
\end{equation}

In standard generative modeling \cite{liu2023flow, EsserKBEMSLLSBP24}, a commonly used path is the linear Optimal Transport, which describes the process from Gaussian noise towards the target distribution through a linear conditional probability path.
\begin{align}
\mu_t &= t x_1 \nonumber \\ 
\sigma_t &= 1 - (1 - \sigma_{\min}) t \label{eq:ot_path}
\end{align}
with $p_t(x|x_1)$:
\begin{equation}
p_t(x|x_1) = \mathcal{N}\left(x \mid t x_1, \sigma_t^2 I\right),
\end{equation}
and corresponding flow and velocity field:
\begin{equation}
\psi_t(x) = t x_1 + \left(1 - (1 - \sigma_{\min}) t\right) x,
\label{eq:ot_flow}
\end{equation}
\begin{equation}
u_t(x|x_1) = \frac{x_1 - (1 - \sigma_{\min}) x}{1 - (1 - \sigma_{\min}) t}.
\label{eq:ot_velocity}
\end{equation}

While effective for general data generation, this "noise-to-data" paradigm is not well-suited for physical systems like weather forecasting, where a future state evolves from a known previous state, not from random noise.
To better align the learning process with the physics of our problem, we introduce a novel probability path \cite{lim2025a} which we term Dynamic Transport. This path is specifically designed for forecasting dynamical systems by initializing the transport from the prior atmospheric state instead of from Gaussian noise. The path is defined as:

\begin{equation}
p_t(x|x_1) = \mathcal{N}(x \mid t x_1 + (1 - t) x_0, \sigma_{\min}^2 I),
\end{equation}

with

\begin{align}
\mu_t &= t x_1 + (1 - t) x_0 \nonumber \\
\sigma_t &= \sigma_{\min} \label{eq:dynamic_path}
\end{align}

and corresponding flow and velocity field:
\begin{equation}
\psi_t(x) = t x_1 + (1 - t) x_0 + \sigma_{\min} x,
\quad
u_t(x|x_1) = x_1 - x_0.
\label{eq:dynamic_flow}
\end{equation}

In our setting, ${x_0}$ and ${x_1}$ correspond to atmospheric fields sampled at 6-hour intervals. By using this Dynamic Transport path and setting the minimum noise level $\sigma_{\min}$ to zero for our deterministic task, the model learns the velocity field equivalent to the finite difference between the two states. The key distinction of our approach lies in this prior-informed initialization  (see Supplementary Tables 2 for a comparison between Optimal Transport and Dynamic Transport), which provides a more direct objective for modeling atmospheric dynamics.

\begin{figure}[htbp]
  \vskip 0.2in
  \begin{center}
    \includegraphics[width=0.99\textwidth]{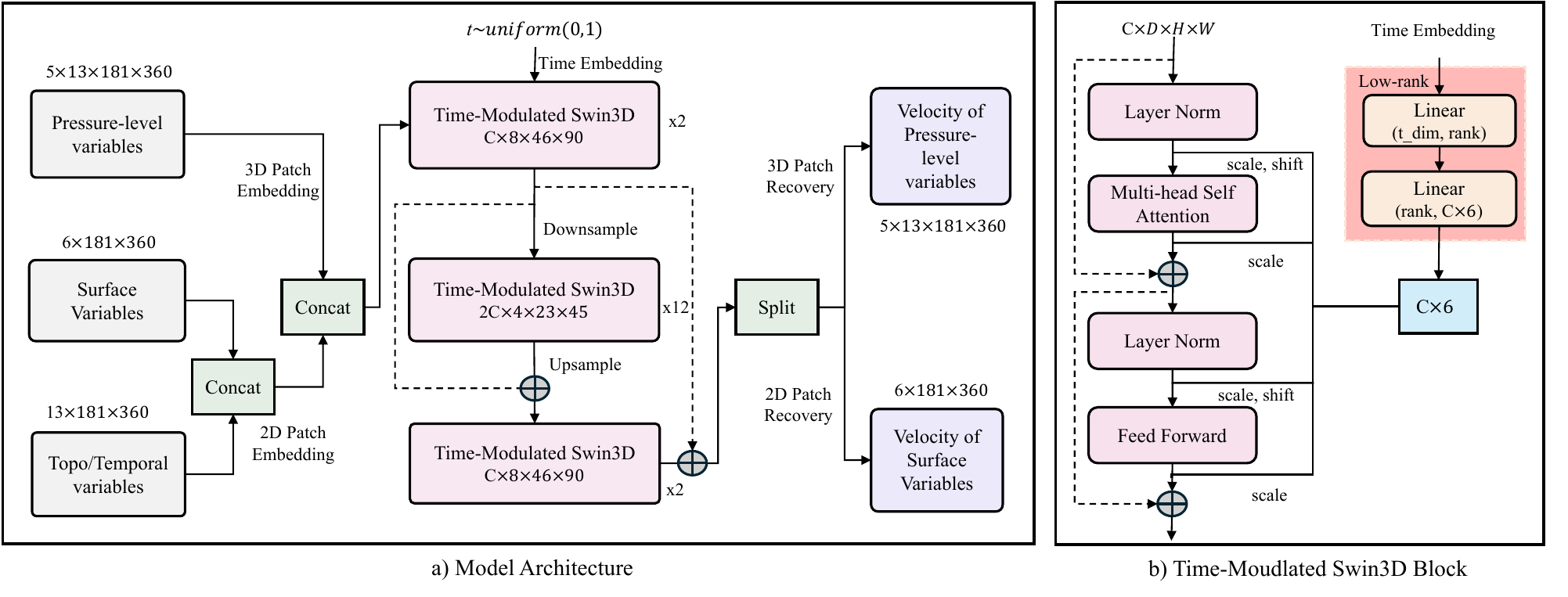}
    \caption{\textbf{Model architecture.} \textbf{a} the overall architecture of the velocity model. \textbf{b} the low-rank time-modulated 3D Swin Transformer block.}
    \label{fig:model}
  \end{center}
  \vskip -0.2in
\end{figure}

\subsubsection{Time-modulated 3D Swin Transformer Architecture}
\label{appendix_arch}
The velocity model consists of three main components: patch embedding, time-modulated 3D Swin Transformer blocks, and patch recovery (see Fig.~\ref{fig:model}a).  

The input data consists of 6 surface variables ($6 \times 181 \times 360$) and 5 pressure-level variables across 13 pressure levels ($5 \times 13 \times 181 \times 360$), along with 11 conditioning features, the binary land–sea mask, geopotential at the surface, soil type, the sine and cosine of latitude and longitude, the sine and cosine of the local time of day and the year progress (all normalized to $[0,1)$). 
Pressure-level variables are processed via a 3D convolutional patch embedding, producing a latent feature map of size $256 \times 7 \times 46 \times 90$. Surface variables and additional conditioning features are concatenated and embedded jointly using a 2D convolutional patch embedding, generating a feature map of size $256 \times 1 \times 46 \times 90$. 
The two embeddings are concatenated along the channel dimension, yielding a combined feature map of size $256 \times 8 \times 46 \times 90$, which is then fed into the subsequent 3D Swin Transformer blocks.

The embedded patches are processed by three consecutive layers containing 2, 12, and 2 3D Swin Transformer blocks \cite{Liu2021}, respectively. A downsampling operation is applied after the first layer, reducing the spatial size while doubling the number of channels to $512 \times 4 \times 23 \times 45$. A corresponding upsampling step is applied before the final layer, restoring the feature map to $256 \times 8 \times 46 \times 90$. Residual connections are applied such that the output of the first layer is added to the outputs of the second and third layers \cite{He2016Deep}, allowing the model to retain low-level features across all subsequent layers and improving gradient flow. 

Each 3D Swin Transformer block incorporates a time modulation module to provide temporal conditioning, where the time is first mapped into a sinusoidal embedding, which is then transformed by a small feed-forward network to modulate the attention mechanism adaptively (see Fig.~\ref{fig:model}b). This design is inspired by the Adaptive-Zero Layer Normalization (adaLN-Zero) \cite{Peebles2023}. Specifically, the normalized feature representation $\tilde{\mathbf{h}}$, obtained by applying LayerNorm \cite{ba2016layernormalization} to $\mathbf{h}$, is modulated by time-dependent scale $\gamma(t)$ and shift $\beta(t)$ parameters:
\begin{equation}
\mathrm{adaLN}(\mathbf{h}; t) = \gamma(t) \odot \tilde{\mathbf{h}} + \beta(t)
\end{equation}

The original adaLN-Zero mapping from the time embedding dimension $t_{\mathrm{dim}}$ to the hidden dimension $6C$ is full-rank, substantially increasing the parameter count. To reduce memory and computational costs, we replace this full-rank projection with a low-rank decomposition:
\begin{equation}
M_{t_{\mathrm{dim}} \times 6C}  \approx U_{t_{\mathrm{dim}} \times r} V_{r \times 6C} 
\end{equation}
where $r=32$ in this study. This reduces the total parameter count from 54.2M to 45.7M, enabling a lighter and more efficient model without compromising performance (see Supplementary Notes 1.2 for details).

Finally, the latent feature map is split into two components of sizes $256 \times 1 \times 46 \times 90$ and $256 \times 7 \times 46 \times 90$, corresponding to surface and pressure-level features, respectively. Each component is then projected back to the original variable dimensions through a linear projection, restoring the spatial resolution and channel dimension for the final velocity field output.

\subsection{Training Details}

\subsubsection{Training Objective}

In the pre-training stage, the velocity model is trained on ERA5 pairs $\{X_k, X_{k+6}\}$ sampled at 6-hour intervals (00/06/12/18:00 UTC) by minimizing the loss:
\begin{equation}
\mathcal{L}_{\text{stage1}}(\theta)
= \frac{1}{C\,H\,W} 
\sum_{s=1}^{C} \sum_{i=1}^{H} \sum_{j=1}^{W} 
w_{\mathrm{lat}}(i)\, w_{\mathrm{lev}}(\ell_s)\, w_{\mathrm{var}}(s) \;
\big[ v_\theta({x}_t, t, c)_{s,i,j} - (X_{k+6} - X_k)_{s,i,j} \big]^2,
\label{eq:stage1-loss}
\end{equation}
where
\begin{itemize}
    \item $w_{\mathrm{lat}}(i)$ — proportional to the surface area of the grid cell at latitude $\varphi_i$, normalized to unit mean, ensuring uniform spatial weighting.
    \item $w_{\mathrm{lev}}(\ell_s)$ — down-weights higher atmospheric levels where air density is lower.
    \item $w_{\mathrm{var}}(s)$ — the inverse variance of the temporal difference $(X_{k+6} - X_k)_s$, balancing variables with different magnitudes of change.
\end{itemize}

The intermediate state ${x}_t$ is obtained by linear interpolation between $X_k$ and $X_{k+6}$:
\begin{equation}
{x}_t = (1-t){X}_k + t{X}_{k+6}, \quad t \in (0,1).
\end{equation}

In the second stage, the model is fine-tuned with ERA5 hourly data to refine short-term evolution.  
Given forecasts $\hat{X}_{\tau}$ and corresponding ground truth $X_{\tau}$ at lead time $\tau$, the loss is defined as
\begin{equation}
\mathcal{L}_{\text{stage2}}(\theta)
= \sum_{\tau=1}^{T}
w_{\tau}
\left(
\frac{1}{C\,H\,W}
\sum_{s=1}^{C} \sum_{j=1}^{H} \sum_{k=1}^{W}
w_{\mathrm{lat}}(j)\, w_{\mathrm{lev}}(\ell_s)\, w_{\mathrm{var}}(s) 
\big[ (\hat{X}_\tau)_{s,j,k} - (X_\tau)_{s,j,k} \big]^2
\right),
\label{eq:stage2-loss}
\end{equation}
where
\begin{equation}
w_{\tau} = \left(1 + \frac{\tau}{24}\right)^{-1/2}.
\end{equation}

The weighting $w_{\tau}$ rescales each trajectory to account for the near-linear increase in expected error variance with lead time~$\tau$, reducing the dominance of long-lead errors and encouraging stable short-term dynamics.

\subsubsection{Model Training}
We train both Pangu-Weather and FlowCast-ODE using the same overall training framework, implemented in PyTorch \cite{paszke2019pytorch}. 
For Pangu-Weather, we follow the implementation released in NVIDIA Modulus\footnote{\url{https://github.com/NVIDIA/modulus}}, 
training a 24-hour model on ERA5 data from 1980–2019 for 50 epochs. 
The optimizer is AdamW \cite{Kingma2015, Loshchilov2019} with weight decay of 0.1 and parameters (\(\beta_1 = 0.9\), \(\beta_2 = 0.95\)), and the learning rate is scheduled using cosine annealing starting from $3\times 10^{-4}$.  

FlowCast-ODE training consists of two stages. 
In Stage~1, we train the 6-hour velocity model from scratch for 50 epochs using 1980–2019 ERA5 data. 
In Stage~2, we fine-tune the Stage~1 model for iterative hourly forecasting, 
performing fine-tuning runs for 6, 12, and 18 autoregressive (AR) steps. 
Each fine-tuning run uses 2000–2019 data and is trained for 5, 2, and 2 epochs, respectively, 
with learning rates $1\times 10^{-6}$, $1\times 10^{-6}$, and $3\times 10^{-7}$.  
To reduce memory usage in Stage~2, gradient checkpointing is applied.  

During training, each trajectory is rescaled to account for the near-linear increase in expected error variance with lead time~$\tau$, 
using the scaling factor $(1+(\tau/24))^{-1/2}$. Additionally, we apply an Exponential Moving Average (EMA) of model parameters with a decay rate of 0.999 \cite{Karras2023AnalyzingAI} for FlowCast-ODE, following common practice in flow matching. 
A full list of hyperparameters for models is given in Supplementary Table 3.

Training the 24-hour Pangu-Weather model required approximately 70 hours using 4 NVIDIA A100 40GB GPUs. 
For FlowCast-ODE, the first-stage training on 6-hour intervals took about 66 hours, 
while the subsequent fine-tuning stage on hourly data required an additional 43 hours on the same hardware.

\subsection{Evaluation Methods}

For variable $X_s$ at forecast time step $k$, the latitude-weighted RMSE is
\begin{equation}
\mathrm{RMSE}_j(k)
= \sqrt{\frac{1}{N_{\mathrm{lat}}N_{\mathrm{lon}}}
\sum_{i=1}^{N_{\mathrm{lat}}}\sum_{j=1}^{N_{\mathrm{lon}}}
w_{\mathrm{lat}}(i)\,\big(\hat{X}_s(i,j,k)_s(i,j,k)-X_s(i,j,k)\big)^2},
\label{eq:rmse_weighted}
\end{equation}
where $\hat{X}_s(i,j,k)$ and ${X}_s(i,j,k)$ denote the forecast and reference (ERA5) values at grid point $(i,j)$ and time step $k$.
The latitude weight $w_{\mathrm{lat}}(i)$ is proportional to the surface area of the grid cell at latitude $\phi_i$ and normalized to have unit mean over the grid.

We evaluated the blurriness of forecasts using the power spectrum, implemented via the ZonalEnergySpectrum class from WeatherBench 2\footnote{\url{https://github.com/google-research/weatherbench2/blob/main/weatherbench2/derived_variables.py}}.

\backmatter

\bmhead{Acknowledgements}
 We would like to acknowledge helpful comments and suggestions from Shikai Fang that strengthened the presentation of this work. We extend our appreciation to ECMWF and WeatherBench2 for providing the ERA5 dataset, and to the NOAA National Centers for Environmental Information for the IBTrACS dataset. Shuo Wang acknowledges the financial support by China Scholarship Council (CSC) Grant No. 202106040117.

\section*{Data availability}
We used a subset of the 1° resolution ERA5 reanalysis dataset to train and evaluate FlowCast-ODE, accessed from the WeatherBench2 Google Cloud bucket \url{https://console.cloud.google.com/storage/browser/weatherbench2/datasets/era5}. To compare tropical cyclone tracks with the 0.25° Pangu-Weather model, 6-hour interval forecasts were obtained from \url{https://console.cloud.google.com/storage/browser/weatherbench2/datasets/pangu}.  Hourly Pangu-Weather forecasts were generated using the official Pangu-Weather code and model weights \url{https://github.com/198808xc/Pangu-Weather} to assess kinetic and internal energy evolution. Observed tropical cyclone tracks were sourced from the International Best Track Archive for Climate Stewardship (IBTrACS) \url{https://www.ncei.noaa.gov/products/international-best-track-archive}.

\section*{Code availability}
The FlowCast-ODE code base was developed using PyTorch Lightning and Hydra, based on the template, \url{https://github.com/ashleve/lightning-hydra-template}. Pangu-Weather follows the implementation released in NVIDIA Modulus, \url{https://github.com/NVIDIA/physicsnemo}. The calculation of power spectrum relied on the ZonalEnergySpectrum class from WeatherBench2, \url{https://github.com/google-research/weatherbench2/blob/main/weatherbench2/derived_variables.py}. The energy computation follows the implementation available at \url{https://github.com/NCAR/CREDIT-physics-run/blob/main/physics/DEV01_global_energy_conservation_plevel.ipynb}. The trained FlowCast-ODE model weights and associated code will be made publicly available upon publication.

\section*{Author contribution}
S.H. prepared the datasets, designed and trained the models, performed evaluations, created figures, wrote and revised the manuscript.  
Y.Z. and X.Y. managed and oversaw the project.  
Q.M. and H.L. improved the model design.  
H.L., Y.Z., Q.M. and S.W. revised the manuscript.

\section*{Competing Interests}
The authors declare no financial or non-financial competing interests related to this work.

\bibliography{sn-bibliography}% common bib file

\setcounter{figure}{0} 
\renewcommand{\thefigure}{\arabic{figure}}
\renewcommand\figurename{Supplementary Figure}% defined as per springer style 
\renewcommand\tablename{Supplementary Table}%
\renewcommand{\thesection}{\arabic{section}}
\setcounter{section}{0}

\clearpage  

\section*{Supplementary information}
\section{Supplementary Notes}
\subsection{Assimilation Window Discontinuities}
\label{appendix_discontinuities}
The ERA5 reanalysis data are produced by ECMWF’s 4D-Var data assimilation system, which assimilates observations over 12-hour windows (09:00–21:00 UTC and 21:00–09:00 UTC). When examining RMSE evolution over a 120-hour forecast horizon, we observed sudden increases at the assimilation window boundaries, suggesting possible inconsistencies across the temporal edges. To investigate this further, we computed four vertically integrated components of atmospheric energy, which allowed us to identify temporal discontinuities in the ERA5 dataset. 

The total energy of the air column is defined as (\cite{mayer2021}):

\begin{equation}
E = \frac{1}{g} \int_{0}^{p_s} \left[(1-q) c_v T_c + L_v(T_c) q + \phi + k \right] dp,
\end{equation}

with the terms representing:

\begin{itemize}
\item Internal energy: $c_v T$, where $c_v$ is the specific heat capacity of air at constant volume and $T_c$ is the air temperature in Celsius.
\item Latent heat energy: $L_v(T_c) q$, with the temperature-dependent latent heat of vaporization $L_v(T_c)$, is defined according to Part IV of the IFS documentation \cite{IFSPartIV}, and $q$ is the specific humidity. 
\item Potential energy: $\phi$, the geopotential.
\item Kinetic energy: $k = \frac{1}{2}(u^2 + v^2)$, defined in terms of the horizontal velocity components of the air parcel.
\end{itemize}

Here, $g$ is the gravitational acceleration. Visualizations of these energy components are presented in Supplementary Figure~\ref{fig:energy}.

\subsection{Low-Rank adaLN-Zero}
\begin{figure}[htbp]
  \centering
  \includegraphics[width=0.88\textwidth]{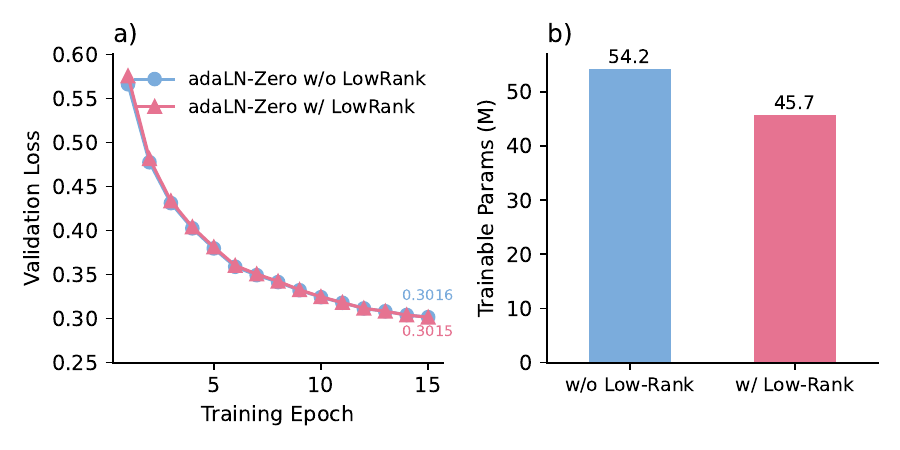}
  \caption{\textbf{Validation loss curves (a) and model parameters (b) for FlowCast-ODE using adaLN-Zero temporal modulation with and without Low-Rank decomposition.} Both models are trained on ERA5 data from 2010–2019, and the curves are shown across training epochs.}
  \label{fig:loss}
  \vspace{-5pt}
\end{figure}

Supplementary Figure~\ref{fig:loss} shows validation loss curves for FlowCast-ODE with adaLN-Zero modulation, comparing the full-rank (54.2M parameters) and Low-Rank (45.7M parameters) variants. The Low-Rank model achieves comparable, and slightly better loss throughout training, demonstrating that the Low-Rank design reduces model size without sacrificing predictive performance.

\subsection{Typhoon Tracking Algorithm}
\label{appendix_typhoon_track}
To track tropical cyclone centers, we employ a classical algorithm that identifies the local minimum of MSLP near a prescribed initial cyclone-eye location at 6-hour forecast intervals \cite{white2005}. A detected MSLP minimum is considered a tropical cyclone only if it satisfies the following dynamical and thermodynamical criteria:
\begin{itemize}
    \item Maximum 850 hPa relative vorticity exceeding $5 \times 10^{-5}\text{s}^{-1}$ within a 278 km radius in the Northern Hemisphere (or below $-5 \times 10^{-5}\text{s}^{-1}$ in the Southern Hemisphere);
    \item Maximum 850–200 hPa thickness within a 278 km radius for extratropical systems;
    \item Maximum 10 m wind speed exceeding $8\text{m s}^{-1}$ within a 278 km radius over land.
\end{itemize}

Once the cyclone eye is confirmed, the tracker iteratively searches for subsequent positions within a 445 km vicinity. The procedure terminates when no local MSLP minimum meets the above conditions.

\section{Supplementary Tables}

\begin{table}[htbp]
\centering
\caption{List of variables used in the experiments}
\label{tab:variables}
\begin{tabular}{llll}
\toprule
\textbf{Type} & \textbf{Variable name} & \textbf{Short name} & \textbf{Unit} \\
\midrule
Atmospheric & Geopotential & Z & m$^2$ s$^{-2}$ \\
Atmospheric & Specific humidity & Q & kg kg$^{-1}$ \\
Atmospheric & Temperature & T & K \\
Atmospheric & U component of wind & U & m s$^{-1}$ \\
Atmospheric & V component of wind & V & m s$^{-1}$ \\
\midrule
Single & 10-m u wind component & U10M & m s$^{-1}$ \\
Single & 10-m v wind component & V10M & m s$^{-1}$ \\
Single & 2-m temperature & T2M & K \\
Single & 2-m dewpoint temperature & TD2M & K \\
Single & Mean sea level pressure & MSLP & Pa \\
Single & Total Precipitation & TP & kg m$^{3}$ \\
\midrule
Static & Geopotential at surface & Z & m$^2$ s$^{-2}$ \\
Static & Land-sea mask & -- & -- \\
Static & Soil type & -- & -- \\
Static & Latitude & -- & -- \\
Static & Longitude & -- & -- \\
\midrule
Clock & Local time of day & -- & -- \\
Clock & Elapsed year progress & -- & -- \\
\bottomrule
\end{tabular}
\end{table}

\begin{table}[!t]
\caption{Summary of Optimal Transport and Dynamic Transport}
\centering
\begin{small}
\setlength{\tabcolsep}{3pt} % Moderate column spacing
\renewcommand{\arraystretch}{1.4} % Row spacing
\begin{tabular}{c c c} % Three centered columns
\toprule
 & \textbf{Optimal Transport} & \textbf{Dynamic Transport} \\
\midrule
$\mu_t$ & $t x_1$ & $t x_1 + (1-t)x_0$ \\
$\sigma_t$ & $1 - (1 - \sigma_{\min})t$ & $\sigma_{\min}$ \\
$p_t(x|x_1)$ & $\mathcal{N}(x \mid t x_1, \sigma_t^2 I)$ & $\mathcal{N}(x \mid t x_1 + (1-t)x_0, \sigma_{\min}^2 I)$ \\
$\psi_t(x)$ & $t x_1 + (1 - (1 - \sigma_{\min})t)x$ & $t x_1 + (1-t)x_0 + \sigma_{\min}x$ \\
$u_t(x|x_1)$ & $\dfrac{x_1 - (1 - \sigma_{\min})x}{1 - (1 - \sigma_{\min})t}$ & $x_1 - x_0$ \\
$u_t(\psi_t(x_0)|x_1)$ & $x_1 - (1 - \sigma_{\min})x_0$ & $x_1 - x_0$ \\
\bottomrule
\end{tabular}
\end{small}
\label{tab:optimal_dyn_transport}
\end{table}

\begin{table}[htbp]
\centering
\caption{A full list of hyper-parameters used for the training of Pangu-Weather and FlowCast-ODE.}
\begin{tabular}{lcccc}
\toprule
Hyper-parameter & Pangu-Weather & FlowCast-ODE Stage 1 & FlowCast-ODE Stage 2 \\
\midrule
Forecast interval & 24 h & 6 h & 1 h \\
Epochs & 50 & 50 & 5 / 2 / 2 \\
Batch size & 16 & 16 & 4 / 4 / 4 \\
Number of AR steps & 1 & 1 & 6 / 12 / 18 \\
Training data & 1980–2019 & 1980–2019 & 2000–2019 \\
Optimizer & AdamW & AdamW & AdamW \\
Weight decay & 0.1 & 0.1 & 0.1 \\
LR decay schedule & Cosine & Cosine & Constant \\
Peak LR & $3e-4$ & $3e-4$ & $1e-6$ / $1e-6$ / $3e-7$ \\
Gradient checkpointing & No & No & Yes \\
\bottomrule
\end{tabular}
\label{tab:training-hparams}
\end{table}

\clearpage

\section{Supplementary Figures}
\begin{figure*}[htbp]
  \vskip 0.2in
  \begin{center}
    \includegraphics[width=0.99\textwidth]{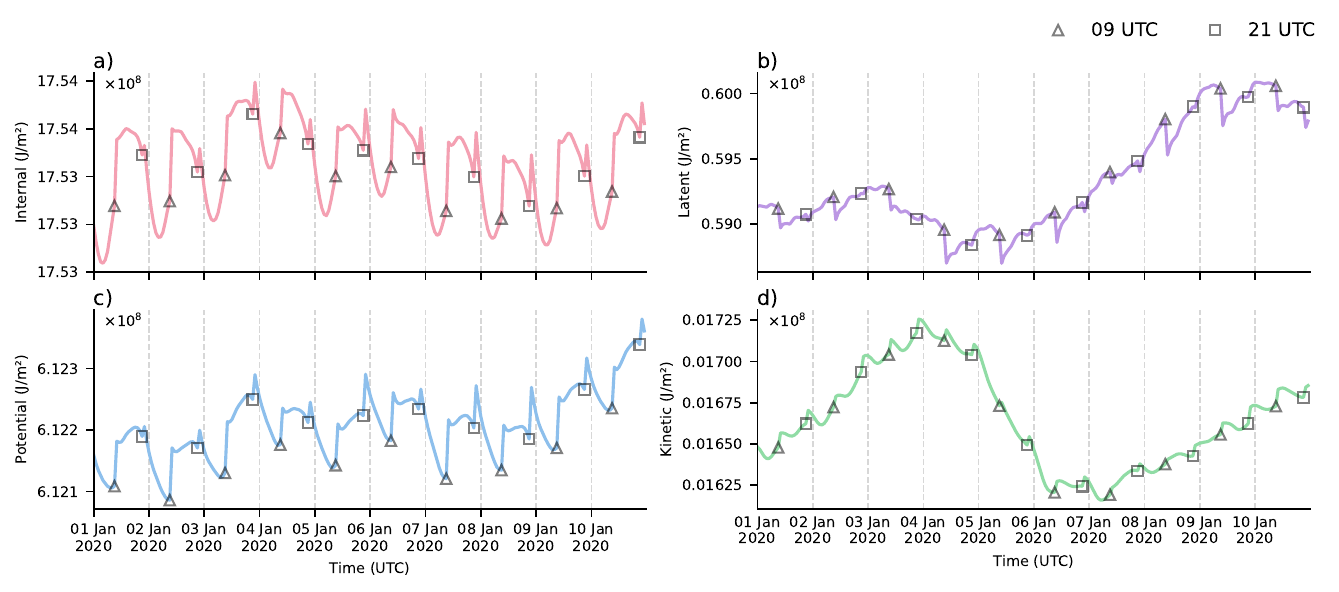}
    \caption{\textbf{Time series of four atmospheric energy components derived from ERA5 data.} \textbf{a} internal energy, \textbf{b} latent heat energy, \textbf{c} potential energy, and \textbf{d} kinetic energy. Triangles indicate values at 09 UTC, while square boxes represent values at 21 UTC, highlighting the discontinuities in the dataset.}
    \label{fig:energy}
  \end{center}
  \vskip -0.3in
\end{figure*}

\end{document}